# The Language of Search

**Jinbo Huang** JINBO.HUANG@NICTA.COM.AU
*Logic and Computation Program*
*National ICT Australia*

**Adnan Darwiche** DARWICHE@CS.UCLA.EDU
*Computer Science Department*
*University of California, Los Angeles*

## Abstract

This paper is concerned with a class of algorithms that perform exhaustive search on propositional knowledge bases. We show that each of these algorithms defines and generates a propositional language. Specifically, we show that the trace of a search can be interpreted as a combinational circuit, and a search algorithm then defines a propositional language consisting of circuits that are generated across all possible executions of the algorithm. In particular, we show that several versions of exhaustive DPLL search correspond to such well-known languages as FBDD, OBDD, and a precisely-defined subset of d-DNNF. By thus mapping search algorithms to propositional languages, we provide a uniform and practical framework in which successful search techniques can be harnessed for compilation of knowledge into various languages of interest, and a new methodology whereby the power and limitations of search algorithms can be understood by looking up the tractability and succinctness of the corresponding propositional languages.

## 1. Introduction

Systematic search algorithms lie at the core of a wide range of automated reasoning systems, and many of them are based on a procedure where branches of a node in the search tree are generated by splitting on the possible values of a chosen variable. One prototypical example is the DPLL algorithm (Davis, Logemann, & Loveland, 1962) for propositional satisfiability (SAT). Given a propositional formula, the problem of SAT is to determine whether the formula has a *satisfying assignment*—an assignment of Boolean values (0 and 1) to the variables under which the formula evaluates to 1. For example, $(x_1 = 0, x_2 = 1, x_3 = 0)$ is a satisfying assignment for the following formula: $(x_1 \vee x_2) \wedge (x_1 \vee \neg x_2 \vee \neg x_3) \wedge (\neg x_1 \vee x_2 \vee \neg x_3)$. To determine the satisfiability of a given formula $\Delta$, DPLL chooses a variable $x$ from the formula, recursively determines whether $\Delta$ is satisfiable in case $x$ is set to 0, and in case $x$ is set to 1, and declares $\Delta$ satisfiable precisely when at least one of the two cases results in a positive answer. In effect, this algorithm performs a systematic search in the space of variable assignments and terminates either on finding a satisfying assignment, or on realizing that no such assignment exists.

Despite its simplicity, DPLL has long remained the basis of most SAT solvers that employ systematic search (Berre & Simon, 2005), and finds natural counterparts in the more general constraint satisfaction problems, where variables are not restricted to the Boolean domain. Several decades of sustained research has greatly enhanced the efficiency and scalability of DPLL-based search algorithms, and today they are routinely used to solve practical problems with several million variables (Zhang & Malik, 2002). These algorithms have been so successful, indeed, that it has become a recent trend in such areas as formal verification to modify them to produce *all solutions* of a propositional formula (McMillan, 2002; Chauhan, Clarke, & Kroening, 2003; Grumberg, Schuster, & Yadgar, 2004), as an alternative to the traditional practice (McMillan, 1993) of converting the formula into an ordered binary decision diagram (OBDD) (Bryant, 1986). The modifications thus involved are in such a form that the DPLL search will not terminate on finding the first solution, but is extended to exhaust the whole search space. For the same formula shown earlier, an example output of such an *exhaustive search* could be the following set of three solutions $\{(x_1 = 0, x_2 = 1, x_3 = 0), (x_1 = 1, x_2 = 0, x_3 = 0), (x_1 = 1, x_2 = 1)\}$. Note that a *solution* here is defined as an assignment of Boolean values to some (possibly all) of the variables that satisfies the formula regardless of the values of the other variables. The last solution in the above set, for example, represents two satisfying assignments as variable $x_3$ is free to assume either value.

Producing all solutions of a propositional formula is, of course, only one of the possible computational tasks for which an exhaustive search is useful. Other such tasks include counting the number of satisfying assignments of a formula, also known as *model counting* (Birnbaum & Lozinskii, 1999; Bayardo & Pehoushek, 2000; Bacchus, Dalmao, & Pitassi, 2003b; Sang, Bacchus, Beame, Kautz, & Pitassi, 2004; Sang, Beame, & Kautz, 2005), and processing certain types of queries on belief and constraint networks (Dechter & Mateescu, 2004b, 2004a).

In this paper we uncover a fundamental connection between this class of exhaustive search algorithms, and a group of propositional languages that have been extensively studied in the field of knowledge compilation (Darwiche & Marquis, 2002). Specifically, we show that an exhaustive search algorithm based on variable splitting, when run on propositional knowledge bases, defines and generates a propositional language in a very precise sense: The *trace* of a single search, when recorded as a graph, can be interpreted as a combinational circuit that is logically equivalent to the propositional knowledge base on which the search has run; the search algorithm itself then defines a propositional language consisting of all circuits that can be generated by legal executions of the algorithm. We show, in particular, that exhaustive DPLL corresponds to the language of FBDD (free binary decision diagrams) (Blum, Chandra, & Wegman, 1980), exhaustive DPLL with fixed variable ordering corresponds to the language of OBDD (ordered binary decision diagrams) (Bryant, 1986), and exhaustive DPLL with decomposition corresponds to a well defined subset of the language of d-DNNF (deterministic decomposable negation normal form) (Darwiche, 2001).

The establishment of this correspondence supplies a bridge between the field of knowledge compilation, and other areas of automated reasoning, including propositional satisfiability, where search algorithms have been extensively studied. In particular, it leads to the following two sets of theoretical and practical benefits.

First, we show that the class of search algorithms we have described can be immediately turned into knowledge compilers for the respective propositional languages they define, by simply recording the trace of the search. This realization provides a uniform and practical framework in which successful techniques developed in the context of search can be directly used for compilation of knowledge into various languages of interest. In particular, we discuss how recent advances in DPLL search, including sophisticated conflict analysis, dependency-directed backtracking, clause learning, new variable ordering heuristics, and data structures for faster constraint propagation, can be harnessed for building efficient practical knowledge compilers.

Second, we show how, by looking up known properties of propositional languages, we are now able to answer at a fundamental level, and in concrete terms, two important questions regarding the power and limitations of the same class of search algorithms: What can these algorithms do? And what can they not? Specifically, we discuss how the tractability of the propositional language defined by a search algorithm illustrates the power of the algorithm, and how the succinctness and the constraints of the language illustrate its limitations.

We complement these discussions by relating our results to previous work on knowledge compilation and a recent body of work centering on the notion of AND/OR search (Dechter & Mateescu, 2004b, 2004a; Marinescu & Dechter, 2005). In particular, we discuss similarities as well as differences between AND/OR search and d-DNNF compilation.

Finally, we present experimental results on implementations of exhaustive search algorithms that define distinct propositional languages. We use these programs to compile a set of propositional formulas into the respective languages, both to demonstrate the practicality of this knowledge compilation framework, and to empirically illustrate the variation of language succinctness in response to the variation of the search strategy.

The remainder of the paper is organized as follows. Section 2 reviews a number of propositional languages concerned in this work and their theoretical roles and relations in knowledge compilation. Section 3 discusses the DPLL search algorithm for propositional satisfiability and its exhaustive extension, and introduces the notion of interpreting the trace of a search as a combinational circuit. Section 4 is a detailed exposition on mapping variants of exhaustive DPLL to FBDD, OBDD, and a subset of d-DNNF, as well as techniques involved in transforming these algorithms into practical knowledge compilers for the respective languages. Section 5 formalizes two fundamental principles relating to the intrinsic power and limitations of a class of exhaustive search algorithms, using recently proposed model counting algorithms as concrete examples. We relate our results to previous work in Section 6, present experimental results in Section 7, and conclude in Section 8. Proofs of all theorems are given in the appendix.

## 2. Propositional Languages and Their Properties

The study of propositional languages (i.e., representations for propositional theories) has been a central subject in knowledge compilation, which is concerned with the task of converting a given knowledge base from one language into another so that certain reasoning tasks become tractable on the compiled knowledge base (Selman & Kautz, 1991; del Val, 1994, 1995; Marquis, 1995; Selman & Kautz, 1996; Cadoli & Donini, 1997; Darwiche & Marquis, 2002; Darwiche, 2002, 2004; Coste-Marquis, Berre, Letombe, & Marquis, 2005). After propositional theories are compiled into the language of OBDD, for example, their equivalence can be tested in time that is polynomial in the sizes of the OBDDs (Meinel & Theobald, 1998), or constant time if the OBDDs use the same variable order (Bryant, 1986). More recent applications of compilation using the language of d-DNNF can be found in the fields of diagnosis (Barrett, 2004, 2005; Huang & Darwiche, 2005a; Elliott & Williams, 2006; Siddiqi & Huang, 2007), planning (Barrett, 2004; Palacios, Bonet, Darwiche, & Geffner, 2005; Huang, 2006; Bonet & Geffner, 2006), probabilistic reasoning (Chavira & Darwiche, 2005; Chavira, Darwiche, & Jaeger, 2006), and query rewrites in databases (Arvelo, Bonet, & Vidal, 2006).

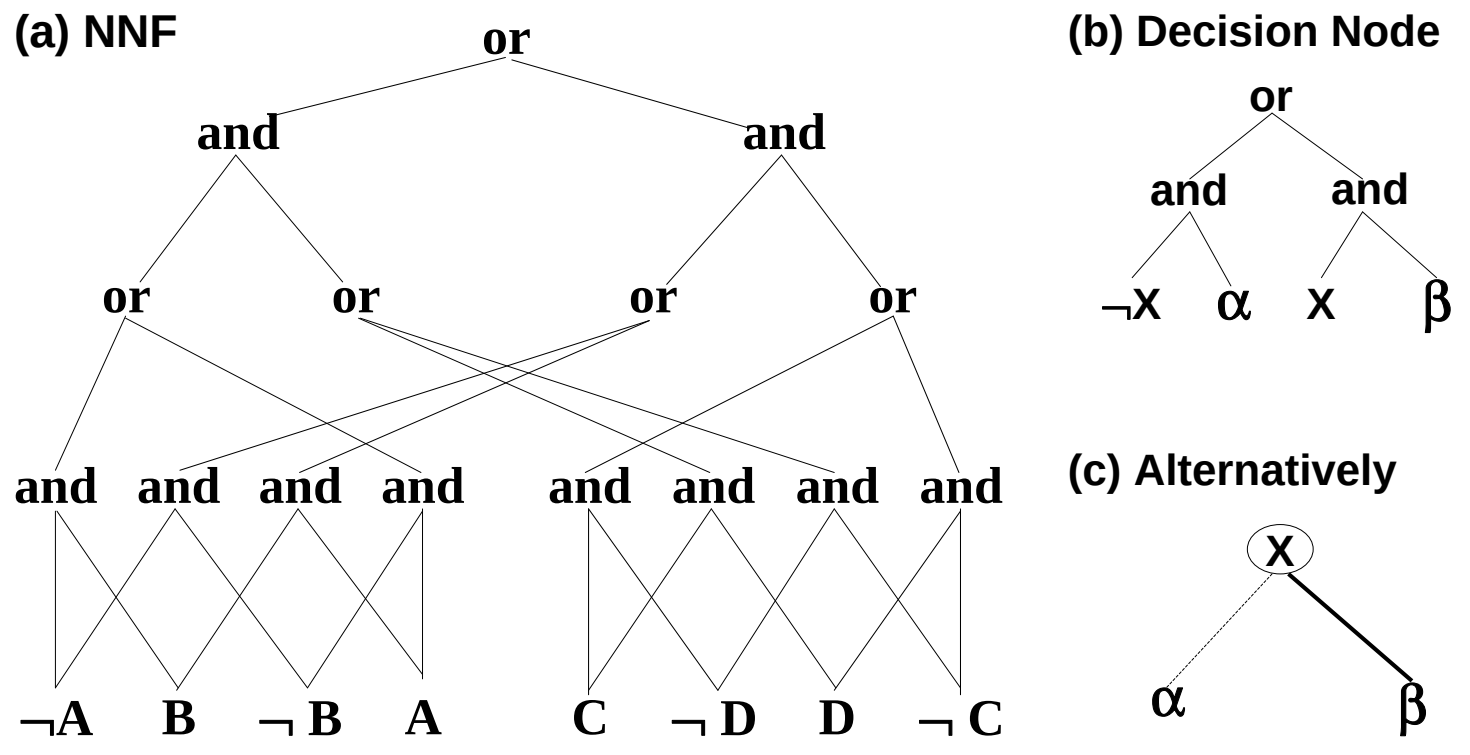

Figure 1: An NNF circuit and a decision node.

In this section we review a set of propositional languages and discuss an established body of results concerning their tractability and succinctness (Darwiche & Marquis, 2002)—several of these languages will resurface in Section 4 as those to which exhaustive search algorithms are mapped, and their properties will prove vital to our formalization in Section 5 of two fundamental principles relating to the power and limitations of exhaustive search algorithms.

## 2.1 Propositional Languages

Following the conventions of Darwiche and Marquis (2002), we consider graph representations of propositional theories, which allow sharing of subformulas for compactness. Specifically, we consider directed acyclic graphs (DAGs) where each internal node is labeled with a conjunction (*and*, $\wedge$) or disjunction (*or*, $\vee$), and each leaf is labeled with a propositional literal or constant ($true/false$, or 1/0). It should be clear that such a DAG is effectively a combinational circuit with and-gates, or-gates, and inverters, where inverters only appear next to inputs (variables), a property that is characteristic of the Negation Normal Form (NNF) (Barwise, 1977). We will hence refer to these DAGs as *NNF circuits* and the set of all such DAGs as the *NNF language*. Figure 1a depicts a propositional theory represented as an NNF circuit. We will next define some interesting subsets of the NNF language.

The popular language of CNF (conjunctive normal form) can now be defined as the subset of NNF that satisfies (i) **flatness**: the height of the DAG is at most two; and (ii) **simple-disjunction**: any disjunction is over leaf nodes only (i.e., is a clause). Similarly, DNF (disjunctive normal form) is the subset of NNF that satisfies flatness and **simple-conjunction**: any conjunction is over leaf nodes only (i.e., is a term).

We consider next a set of nested representations, starting with the DNNF (decomposable negation normal form) language, which is the set of all NNF circuits satisfying **decomposability**: conjuncts of any conjunction share no variables. Our next language, d-DNNF, satisfies both decomposability and **determinism**: disjuncts of any disjunction are pairwise logically inconsistent. The NNF circuit shown in Figure 1a, for example, is in d-DNNF;

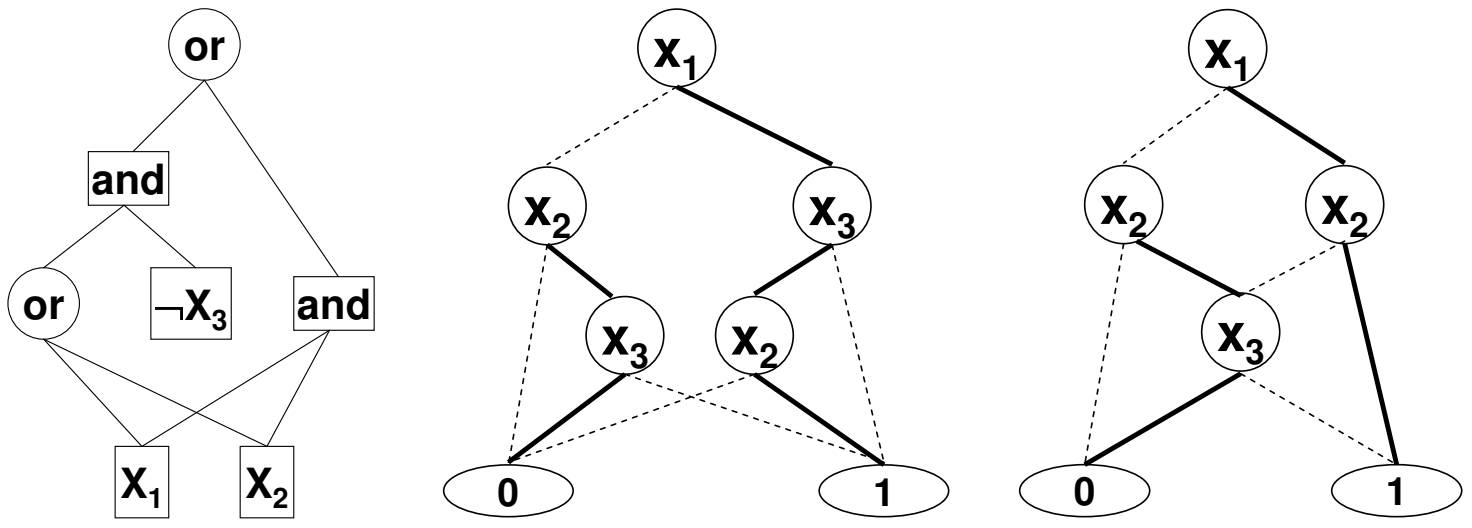

(a) DNNF (b) FBDD (c) OBDD

Figure 2: A circuit in DNNF, FBDD, and OBDD.

that shown in Figure 2a, by contrast, is in DNNF but not in d-DNNF as neither of the two disjunction nodes satisfies determinism.

The FBDD language is the subset of d-DNNF where the root of every circuit is a **decision** node, which is defined recursively as either a constant (0 or 1) or a disjunction in the form of Figure 1b where $X$ is a propositional variable and $\alpha$ and $\beta$ are decision nodes. Note that an equivalent but more compact drawing of a decision node as Figure 1c is widely used in the formal verification literature, where FBDDs are equivalently known as BDDs (binary decision diagrams) that satisfy the **test-once** property: each variable appears at most once on any root-to-sink path (Gergov & Meinel, 1994).[1] See Figure 2b for an FBDD example using this more compact drawing.

The OBDD language is the subset of FBDD where all circuits satisfy the **ordering** property: variables appear in the same order on all root-to-sink paths (Bryant, 1986). See Figure 2c for an OBDD example (using the more compact drawing). For a particular variable order $<$, we also write $\text{OBDD}_<$ to denote the corresponding OBDD subset where all circuits use order $<$.

## 2.2 Succinctness and Tractability of Propositional Languages

Given a choice of languages in which a knowledge base may be represented, one needs to strike a balance between the size of the representation and the support it provides for the reasoning task at hand, as these two properties of a representation often run counter to each other. CNF, for example, is often convenient for compactly encoding a knowledge base since in many applications the behavior of a system can be naturally described as the conjunction of behaviors of its components. However, few typical reasoning tasks can be efficiently carried out on CNF representations. There is no efficient algorithm to determine, for example, whether an arbitrary clause is entailed by a CNF formula. The story changes when the propositional theory is represented in a language known as PI (prime implicates, a subset of CNF): By definition PI supports a linear-time clausal entailment test. The downside is, unfortunately, that PI representations can be exponentially larger than their CNF equivalents in the worst case (Karnaugh, 1953; Forbus & de Kleer, 1993).

We are therefore interested in formally analyzing the *succinctness* and *tractability* of languages, so that given a required reasoning task, we can choose the most succinct language

1. FBDDs are also known as *read-once branching programs* (Wegener, 2000).

that supports the set of necessary operations in polynomial time. The following is a classical definition of succinctness:

**Definition 1.** *(Succinctness) Let $L_1$ and $L_2$ be two subsets of NNF. $L_1$ is at least as succinct as $L_2$, denoted $L_1 \leq L_2$, iff there exists a polynomial $p$ such that for every circuit $\alpha \in L_2$, there exists a logically equivalent circuit $\beta \in L_1$ where $|\beta| \leq p(|\alpha|)$. Here, $|\alpha|$ and $|\beta|$ are the sizes of $\alpha$ and $\beta$, respectively.*

Intuitively, language $L_1$ is at least as succinct as language $L_2$ if given any circuit in $L_2$, there exists a logically equivalent circuit in $L_1$ whose size does not "blow up." One can also define $L_1$ to be *strictly more succinct than* $L_2$, denoted $L_1 < L_2$, if $L_1 \leq L_2$ but $L_2 \not\leq L_1$. The languages we described in Section 2.1 satisfy the following succinctness relations: NNF $<$ DNNF $<$ d-DNNF $<$ FBDD $<$ OBDD, NNF $<$ CNF, and DNNF $<$ DNF (Darwiche & Marquis, 2002). Note, however, that $L_1 \supset L_2$ does not imply $L_1 < L_2$ in general. In other words, imposing conditions on a representation does not necessarily reduce its succinctness. One example is *smoothness*, which requires disjuncts of any disjunction to mention the same set of variables—it is known that this condition, when imposed on d-DNNF, does not reduce its succinctness (Darwiche & Marquis, 2002).

We now turn to the tractability of languages, which refers to the set of polynomial-time operations they support. According to Darwiche and Marquis (2002), one traditionally distinguishes between two types of operations on circuits in a given language: queries and transformations. The difference between the two is that queries return information about circuits, but normally do not change them, while transformations modify circuits to generate new ones (in the same language).

Some of the known results from Darwiche and Marquis (2002) regarding the tractability of languages are summarized in Table 1 (queries) and Table 2 (transformations). The abbreviations in the first row of Table 1 stand for the following eight queries, respectively: **Consistency** (is the formula satisfiable), **Validity** (does the formula evaluate to 1 under all variable assignments), **Clausal Entailment** (does the formula imply a given clause), **Implicant** (is the formula implied by a given term), **Equivalence** (are the two formulas logically equivalent), **Sentential Entailment** (does the one formula imply the other), **Model Counting** (how many satisfying assignments does the formula have), **Model Enumeration** (what are the satisfying assignments of the formula). The abbreviations in the first row of Table 2 stand for the following eight transformations, respectively: **Conditioning** (setting a set of variables to constants), **Forgetting** (existentially quantifying a set of variables), **Single-Variable Forgetting** (existentially quantifying a single variable), **Conjunction** (conjoining a set of circuits), **Bounded Conjunction** (conjoining a bounded number of circuits), **Disjunction** (disjoining a set of circuits), **Bounded Disjunction** (disjoining a bounded number of circuits), **Negation** (negating a circuit).

Interestingly, Table 1 offers one explanation for the popularity of OBDDs in formal verification where efficient equivalence testing, among other things, is often critical. Although more succinct, d-DNNF and FBDD are not known to admit a polynomial-time equivalence test (a polynomial-time probabilistic equivalence test is possible; see Blum et al., 1980; Darwiche & Huang, 2002). Note also that although there is no difference between d-DNNF and FBDD to the extent of this table, the question mark on the equivalence test (EQ) could eventually be resolved differently for the two languages.

Table 1: Polynomial-time queries supported by a language (○ means "not supported unless P=NP" and ? means "we don't know").

| Language | CO | VA | CE | IM | EQ | SE | CT | ME |
|---|---|---|---|---|---|---|---|---|
| NNF | ○ | ○ | ○ | ○ | ○ | ○ | ○ | ○ |
| DNNF | √ | ○ | √ | ○ | ○ | ○ | ○ | √ |
| d-DNNF | √ | √ | √ | √ | ? | ○ | √ | √ |
| BDD | ○ | ○ | ○ | ○ | ○ | ○ | ○ | ○ |
| FBDD | √ | √ | √ | √ | ? | ○ | √ | √ |
| OBDD | √ | √ | √ | √ | √ | ○ | √ | √ |
| OBDD$_<$ | √ | √ | √ | √ | √ | √ | √ | √ |
| DNF | √ | ○ | √ | ○ | ○ | ○ | ○ | √ |
| CNF | ○ | √ | ○ | √ | ○ | ○ | ○ | ○ |

Table 2: Polynomial-time transformations supported by a language (• means "not supported," ○ means "not supported unless P=NP," and ? means " we don't know").

| Language | CD | FO | SFO | $\wedge$C | $\wedge$BC | $\vee$C | $\vee$BC | $\neg$C |
|---|---|---|---|---|---|---|---|---|
| NNF | √ | ○ | √ | √ | √ | √ | √ | √ |
| DNNF | √ | √ | √ | ○ | ○ | √ | √ | ○ |
| d-DNNF | √ | ○ | ○ | ○ | ○ | ○ | ○ | ? |
| BDD | √ | ○ | √ | √ | √ | √ | √ | √ |
| FBDD | √ | • | ○ | • | ○ | • | ○ | √ |
| OBDD | √ | • | √ | • | ○ | • | ○ | √ |
| OBDD$_<$ | √ | • | √ | • | √ | • | √ | √ |
| DNF | √ | √ | √ | • | √ | √ | √ | • |
| CNF | √ | ○ | √ | √ | √ | • | √ | • |

It is also worth pointing out that while tractability with respect to the queries generally improves when the language becomes more restrictive (has more conditions imposed), tractability with respect to the transformations may not. DNNF, for example, supports a subset of the queries that are supported by OBDD according to Table 1, and is therefore less tractable than OBDD from this point of view. However, when it comes to certain transformations, such as the operation of Forgetting (existential quantification), DNNF becomes more tractable than OBDD according to Table 2. The key reason for this shift of advantage is that transformations operate on circuits in a given propositional language, and require the result to be in the same language—this requirement can become a burden for the more restrictive language where more conditions need to be satisfied when the result of the transformation is generated.

### 2.3 A Connection to Be Established

We have summarized and discussed in this section a rich body of known results concerning properties of various propositional languages. These results have previously been presented as a guide for the task of selecting a suitable target compilation language in applications of knowledge compilation. In particular, they suggest that given a reasoning task involving knowledge compilation, one identify the set of operations required for the task, and then select the most succinct target compilation language supporting these operations (Darwiche & Marquis, 2002).

In the following sections of the paper, we wish to establish a fundamental connection between propositional languages with distinct degrees of succinctness and tractability, and exhaustive search algorithms running under distinct sets of constraints. Specifically, we show that the *trace* of an exhaustive search can be interpreted as a circuit representing a compilation of the propositional knowledge base on which the search has run, and the search algorithm itself then defines a propositional language consisting of all its possible traces. This connection will then serve as a bridge between the field of knowledge compilation and other areas of automated reasoning in which search algorithms have been extensively studied, affording two related sets of benefits as follows.

In the first direction, we show how this connection provides a set of practical algorithms for compilation of knowledge into various languages. Specifically, an exhaustive search algorithm can be directly turned into a knowledge compiler by recording the trace of its execution as a graph, and variations on the search algorithm then nicely correspond to compilers for different propositional languages. Such a framework for knowledge compilation provides a significant advantage in that many (past as well as future) advances in search will automatically carry over to knowledge compilation. In particular, we discuss how knowledge compilers can capitalize on several important recent advances in the DPLL search for propositional satisfiability, including sophisticated conflict analysis, dependency-directed backtracking, clause learning, new variable ordering heuristics, and data structures for faster constraint propagation.

In the second direction, we formulate two principles whereby the intrinsic power and limitations of a given exhaustive search algorithm can be understood by identifying the propositional language defined by the search algorithm. Specifically, the tractability of the language illustrates the power (usefulness) of the search algorithm and the succinctness and the constraints of the language illustrates its limitations. Using a group of recently proposed model counters as concrete examples, we show that the search algorithms used by these model counters are powerful enough to support not just a model counting query, but all other queries that are known to be tractable for the language of d-DNNF, such as a probabilistic equivalence test (Darwiche & Huang, 2002). On the other hand, two fundamental limitations can be identified for these same algorithms, as well as other exhaustive search algorithms based on variable splitting, the first in their traces being restricted to a subset of d-DNNF, potentially limiting the efficiency of the search, and the second in their inability to produce traces without determinism, making them overly constrained for compilation of knowledge into more general languages than d-DNNF, such as DNNF.

We now proceed to uncover the connection between search algorithms and propositional languages, starting with a systematic search algorithm for propositional satisfiability, its exhaustive extension, and the notion of the trace of a search.

## 3. Systematic Search for Satisfiability and Its Exhaustive Extension

In this section we introduce the notion of the trace of a systematic exhaustive search, and show how the trace can be interpreted as a circuit, which is logically equivalent to (and hence is a compilation of) the propositional knowledge base on which the search has run. We will do so in the context of systematically searching for satisfying assignments of a propositional formula, a major approach to the problem of propositional satisfiability (SAT) that has come to be known as DPLL (Davis et al., 1962).

### 3.1 DPLL Search and Its Exhaustive Extension

Algorithm 1 is a summary of DPLL for SAT, which takes a propositional formula in CNF, and return 1 (0) precisely when the formula is satisfiable (unsatisfiable). It works by recursively doing a case analysis on the assignment to a selected variable (Line 5): The formula is satisfiable if and only if either case results in a satisfiable subformula (Line 6). The two subformulas are denote as $\Delta|_{x=0}$ and $\Delta|_{x=1}$, which result from replacing all occurrences of $x$ in $\Delta$ with 0 and 1, respectively. In keeping with the rules of Boolean logic, we assume that if a literal becomes or evaluates to 0 as a result of this variable instantiation, it is removed from every clause that contains it; and that if a literal becomes or evaluates to 1, all clauses that contain it are removed. (To facilitate our subsequent discussion of variants of DPLL, we have omitted the use of *unit resolution* from the pseudocode. The programs used in Section 7, however, do employ unit resolution.) In effect, Algorithm 1 performs a search in the space of variable assignments until it finds one that satisfies the given CNF formula or realizes that no satisfying assignments exist.

Now consider extending Algorithm 1 so that it will go through the space of all satisfying assignments—by always exploring both branches of Line 6—rather than terminate on finding the first one. Figure 3a depicts the search tree of this exhaustive version of DPLL, under some particular variable ordering, on the following CNF formula: $(x_1 \vee x_2) \wedge (x_1 \vee \neg x_2 \vee \neg x_3) \wedge (\neg x_1 \vee x_2 \vee \neg x_3)$. Note that in drawing the branches of the search we use a dotted (solid) line to denote setting the variable to 0 (1)—we will also refer to the corresponding child of the search node as the *low* (*high*) child.

**Algorithm 1** DPLL(CNF: $\Delta$): returns satisfiability of $\Delta$

1: **if** there is an empty clause in $\Delta$ **then**
2: return 0
3: **if** there are no variables in $\Delta$ **then**
4: return 1
5: select a variable $x$ of $\Delta$
6: return DPLL($\Delta|_{x=0}$) or DPLL($\Delta|_{x=1}$)

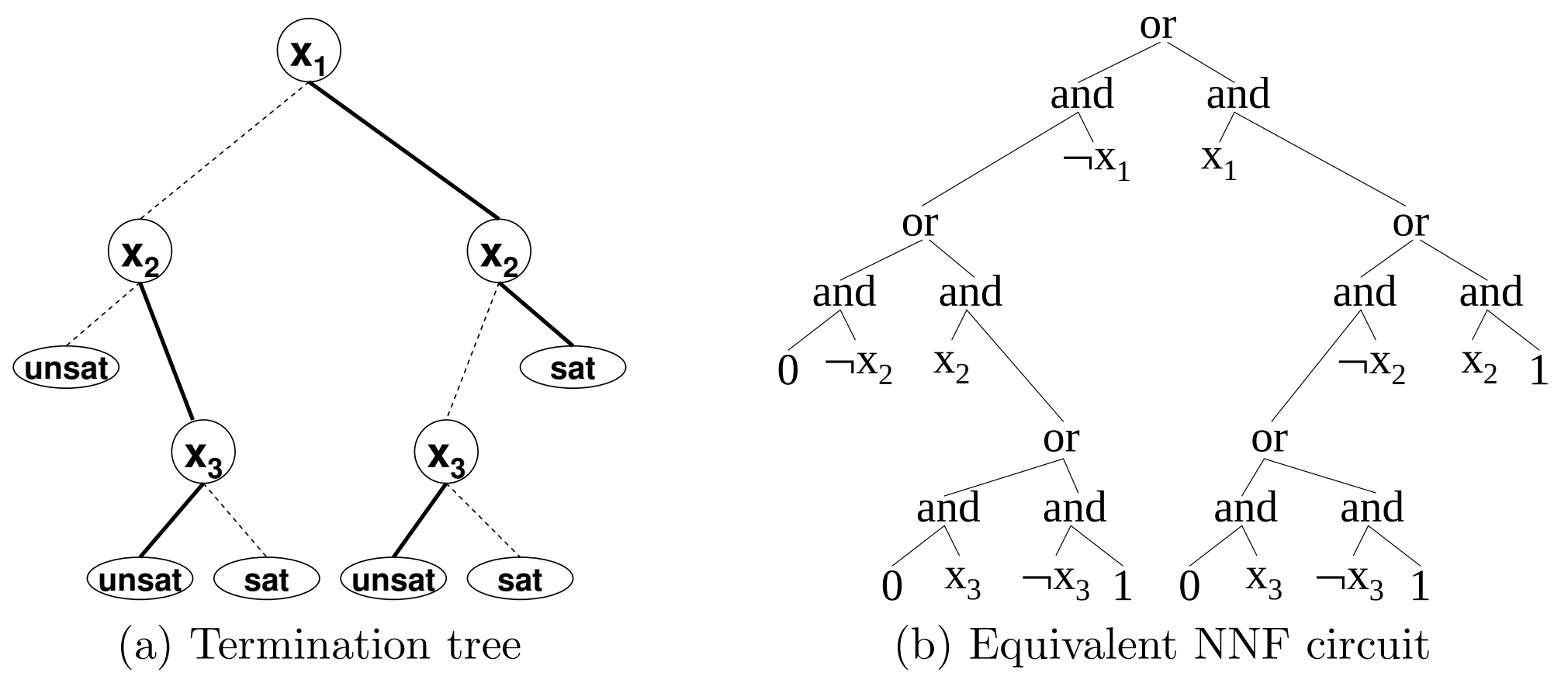

Figure 3: The trace of an exhaustive DPLL search.

The tree depicted in Figure 3a is also known as the *termination tree* of the search, as it captures the set of paths through the search space that have been explored at the termination of the search. In particular, each leaf of the tree labeled with "sat" gives a partial variable assignment that satisfies the propositional formula regardless of the values of any unassigned variables, and the whole tree characterizes precisely the set of all satisfying assignments, which the algorithm has set out to find and succeeded in finding.

### 3.2 The Trace of the Search and the Issue of Redundancy

As we have alluded to earlier, we would like to view such a termination tree as the *trace* of the search, about which we now make two important observations: First, the trace of the search as depicted in Figure 3a can be directly translated into a circuit in NNF as depicted in Figure 3b. All that is involved is to rename "sat/unsat" to "1/0" and invoke the identity between Figure 1b and Figure 1c as we described in Section 2.1. Second, this NNF circuit is logically equivalent to, and hence is a compilation of, the CNF formula on which the search has run. (Note that this notion of trace is different from that used in earlier work to establish the power of DPLL as a proof system for unsatisfiable CNF formulas. For example, in earlier work it was shown that on an unsatisfiable CNF formula, the trace of DPLL can be converted into a tree-like resolution refutation (Urquhart, 1995).)

These two observations imply that exhaustive DPLL is as powerful as a knowledge compiler, as long as one takes the (small) trouble of recording its trace. From the viewpoint of knowledge compilation, however, a search trace recorded in its present form may not be immediately useful, because it will typically have a size proportional to the amount of work done to produce it. Answering even a linear-time query (which may require only a single traversal of the compiled representation) on such a compilation, for example, would be as if one were running the whole search over again.

This problem can be remedied by first realizing that there is quite a bit of redundancy in the search trace we have drawn. In Figure 3a, for example, the two subgraphs whose roots are labeled with $x_3$ are isomorphic to each other and could be merged into one. The same redundancy, of course, is present in the corresponding portions of the NNF circuit shown in Figure 3b.

We can distinguish two levels of dealing with the issue of redundancy in the trace. In the first level, we can remove all redundancy from the trace by reducing it from a tree to a DAG, with repeated applications of the following two rules: (i) Isomorphic nodes (i.e., nodes that have the same label, same low child, and same high child) are merged; (ii) Any node with identical children is deleted and pointers to it are redirected to either one of its children (Bryant, 1986). If we apply these reduction rules to the tree of Figure 3a (again renaming "sat/unsat" to "1/0"), we will get the DAG shown in Figure 2c (in this particular example the second rule does not apply). Note that instead of performing the reduction at the end of the search as these two reduction rules suggest, we can do better by integrating the rules into the trace recording process so that redundant portions of the trace will not be recorded in the first place. This brings us to a technique known as *unique nodes* (Brace, Rudell, & Bryant, 1991; Somenzi, 2004), which we discuss in more detail in the next section.

Removing redundancy in this level ensures that the smallest possible compilation is obtained given a particular execution of the search algorithm; however, it does not improve the time complexity of the search itself. In Figure 3a, for example, the reason why there are two isomorphic subgraphs (rooted at nodes labeled with $x_3$) in the first place is that the search has run into equivalent subproblems from different paths. In the general case these can be nontrivial subproblems, and solving each one of them can be a source of great inefficiency. We therefore refer to the second level of dealing with the issue of redundancy, where we would like to be able to recognize the equivalence of subproblems and avoid carrying out the same computation over and over again. This can be done using the technique of *formula caching* (Majercik & Littman, 1998), which we also discuss in the following section.

## 4. The Language of Search

We have established in Section 3 the notion of interpreting the trace of an exhaustive search as a circuit. In this section we continue our study of these search algorithms by showing how each defines a propositional language consisting of all its possible traces. We will look at three algorithms in particular: (i) the original exhaustive DPLL, (ii) exhaustive DPLL with fixed variable ordering, and (iii) exhaustive DPLL with decomposition. For each algorithm we will discuss the propositional language it defines, the corresponding knowledge compiler it provides, as well as issues regarding the efficiency of the knowledge compiler.

### 4.1 Mapping Exhaustive DPLL to FBDD

We have seen in Section 3 that with the application of reduction rules, the trace of exhaustive DPLL in our example, depicted in Figure 3a, can be stored more compactly as Figure 2c, which is none other than a circuit in the language of FBDD (which happens to be also an OBDD in this case).

We will now formally show that traces of exhaustive DPLL across all possible executions of the algorithm form a propositional language that is precisely the language of (reduced) FBDD as defined in Section 2 (from now on we will assume that circuits in FBDD and OBDD are always given in their reduced form by application of the two reduction rules). In order to do so we will first need a formalism for explicitly recording the trace of the search as a graph. For this purpose we introduce Algorithm 2, which is exactly the exhaustive extension of the original DPLL (Algorithm 1) except that the newly introduced function, **get-node** (given in Algorithm 3), provides a means of recording the trace of the search in the form of a DAG. Specifically, get-node will return a decision node (in the form of Figure 1b) labeled with the first argument, having the second argument as the low child, and having the third argument as the high child (Lines 2&4 have also been modified to return the terminal decision nodes, instead of the Boolean constants). Note that, as we briefly mentioned in Section 3, this algorithm will have its trace recorded directly in the reduced from, instead of producing redundant nodes to be removed later. This is because the two reduction rules are built in by means of a **unique nodes** table, well known in the BDD community (Brace et al., 1991; Somenzi, 2004). Specifically, all nodes created by get-node are stored in a hash table and get-node will not create a new node if (i) the node to be created already exists in the table (that existing node is returned); or (ii) the second and third arguments are the same (either argument is returned). We can now formally state our result as follows:

**Theorem 1.** *DAGs returned by Algorithm 2 form the language of (reduced) FBDD.*

Theorem 1 immediately provides us with a CNF-to-FBDD compiler, which means that as soon as the search finishes, we can answer in polynomial time any query about the propositional theory, as long as that query is known to be tractable for FBDD. According to Table 1, such queries include consistency, validity, clausal entailment, implicant, model counting, and model enumeration. According to Blum et al. (1980), one can then also test the equivalence of two propositional formulas probabilistically in polynomial time after running Algorithm 2 on both. On the other hand, if a propositional theory given to Algorithm 2 is known to have no polynomial-size representation in FBDD, we can also conclude that the algorithm will not be able to finish in polynomial time no matter what variable ordering it uses.

To make Algorithm 2 a practical FBDD compiler, we need to deal with the issue of redundant computation as briefly mentioned in Section 3. The reason is that, despite the use of unique nodes which controls the space complexity, Algorithm 2 still has a time complexity proportional to the size of the tree version of the search trace: Portions of the DAG can end up being explored multiple times. See Figure 4 for an example, where two different instantiations of the first three variables lead to the same subformula, which would then be compiled twice, unnecessarily, by Algorithm 2. To alleviate this problem, one resorts to the technique of *formula caching* (Majercik & Littman, 1998).

Algorithm 4 describes the same exhaustive DPLL search, but now with caching. The result of a recursive call $\text{DPLL}_f(\Delta)$ will be stored in a cache (Line 10) before being returned, indexed by a key (computed on Line 5) identifying $\Delta$; any subsequent call on some $\Delta'$ will

---

**Algorithm 2** $\text{dpll}_f(\text{CNF: } \Delta)$: exhaustive DPLL

1: **if** there is an empty clause in $\Delta$ **then**
2:    return 0-sink
3: **if** there are no variables in $\Delta$ **then**
4:    return 1-sink
5: select variable $x$ of $\Delta$
6: return get-node($x$, $\text{dpll}_f(\Delta|_{x=0})$, $\text{dpll}_f(\Delta|_{x=1})$)

---

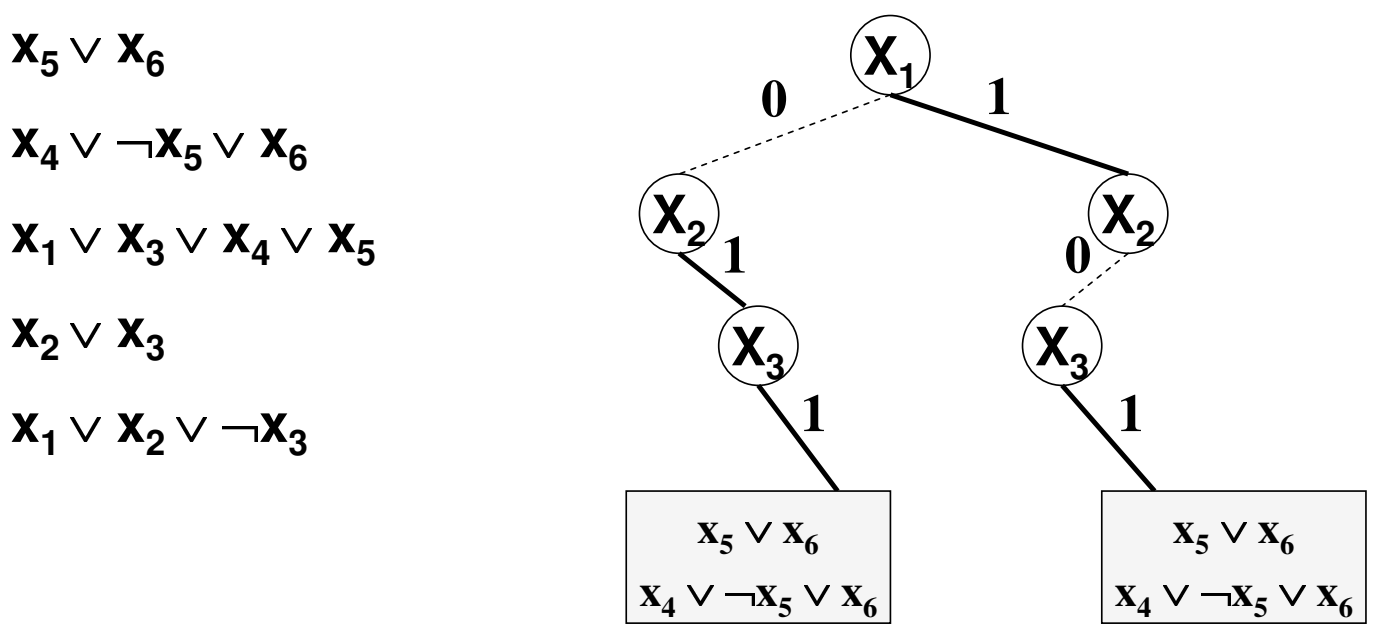

Figure 4: Reaching the same subformula via different paths of the search.

immediately return the existing compilation for $\Delta$ from the cache (Line 7) if $\Delta'$ is found to be equivalent to $\Delta$ (by a key comparison on Line 6). (Note that the introduction of caching does not change the identity of the proposition language defined by the algorithm. In other words, Theorem 1 applies to Algorithm 4 as well.)

In practice, one normally focuses on efficiently recognizing formulas that are *syntactically* identical (i.e., have the same set of clauses). Various methods have been proposed for this purpose in recent years, starting with Majercik and Littman (1998) who used caching for probabilistic planning problems, followed by Darwiche (2002) who proposed a concrete formula caching method in the context of knowledge compilation, then Bacchus, Dalmao, and Pitassi (2003a) and Sang et al. (2004) in the context of model counting, and then Darwiche (2004) and Huang and Darwiche (2005b) who proposed further refinements on the method of Darwiche (2002).

### 4.2 Mapping Exhaustive DPLL with Fixed Variable Ordering to OBDD

Note that in Algorithm 4, DPLL is free to choose any variable on which to branch (Line 8). This corresponds to the use of a dynamic variable ordering heuristic in a typical SAT solver, and is in keeping with the spirit of *free* binary decision diagrams (FBDD).

Not surprisingly, when one switches from dynamic to static variable ordering, the DAGs produced by the algorithm will be restricted to a subset of FBDD. Algorithm 5 implements this change, by taking a particular variable order $\pi$ as a second argument, and making sure that this order is enforced when choosing the next branching point (see Line 8). Across all possible inputs and variable orderings, this algorithm will indeed produce exactly the set of all circuits in the language of (reduced) OBDD:

**Algorithm 3** get-node(int: $i$, BDD: *low*, BDD: *high*)

**if** *low* is the same as *high* **then**
  return *low*
**if** node $(i, low, high)$ exists in *unique-table* **then**
  return *unique-table*$[(i, low, high)]$
$result =$ create-bdd-node$(i, low, high)$
*unique-table*$[(i, low, high)] = result$
return *result*

**Theorem 2.** *DAGs returned by Algorithm 5 form the language of (reduced) OBDD.*

We are therefore provided with a CNF-to-OBDD compiler in Algorithm 5, which means that as soon as the search finishes, we can answer in polynomial time any query about the propositional theory, as long as that query is known to be tractable for OBDD. Most notably, we can now test the equivalence of two propositional formulas deterministically in polynomial time after running Algorithm 5 on both, which we could not with Algorithm 2 or Algorithm 4. On the other hand, if a propositional theory given to Algorithm 5 is known to have no polynomial-size representation in OBDD, such as the *hidden weighted bit function* (Bryant, 1991), we can also conclude that the algorithm will not be able to finish in polynomial time no matter what variable ordering it uses.

To make Algorithm 5 a practical OBDD compiler, we need again to deal with the issue of redundant computation. Naturally, any general formula caching method, such as the ones we described earlier, will be applicable to Algorithm 5. For this more constrained search algorithm, however, a special method is available where shorter cache keys can be used to reduce the cost of their manipulation. The reader is referred to Huang and Darwiche (2005b) for details of this method, which allows one to bound the number of distinct cache keys, therefore providing both a space and a time complexity bound. In particular, with this specific caching scheme in force, the space and time complexity of Algorithm 5 was shown to be exponential only in the *cutwidth* of the given CNF formula. A variant caching scheme allows one to show a parallel complexity in terms of the *pathwidth* (cutwidth and pathwidth are not comparable).

We emphasize here that Algorithm 5 represents a distinct way of OBDD construction, in contrast to the standard method widely adopted in formal verification where one recursively builds OBDDs for components of the system (or propositional formula) to be compiled and combines them using the *Apply* operator (Bryant, 1986). A well-known problem with this latter method is that the intermediate OBDDs that arise in the process can grow so large as to make further manipulation impossible, even when the final result would have a tractable size. Considering that the final OBDD is really all that one is after, Algorithm 5 affords a solution to this problem by building exactly it, no more and no less (although it may do more work than is linear in the OBDD size, both because inconsistent subproblems do not contribute to the OBDD size, and because the caching is not complete). An empirical

**Algorithm 4** $\text{DPLL}_f$(CNF: $\Delta$): exhaustive DPLL with caching

**if** there is an empty clause in $\Delta$ **then**
  return 0-sink
**if** there are no variables in $\Delta$ **then**
  return 1-sink
$key$ = compute-key($\Delta$)
**if** there exists a cache entry ($key, result$) **then**
  return $result$
select variable $x$ of $\Delta$
$result$ = get-node($x$, $\text{DPLL}_f(\Delta|_{x=0})$, $\text{DPLL}_f(\Delta|_{x=1})$)
cache-insert($key$, $result$)
return $result$

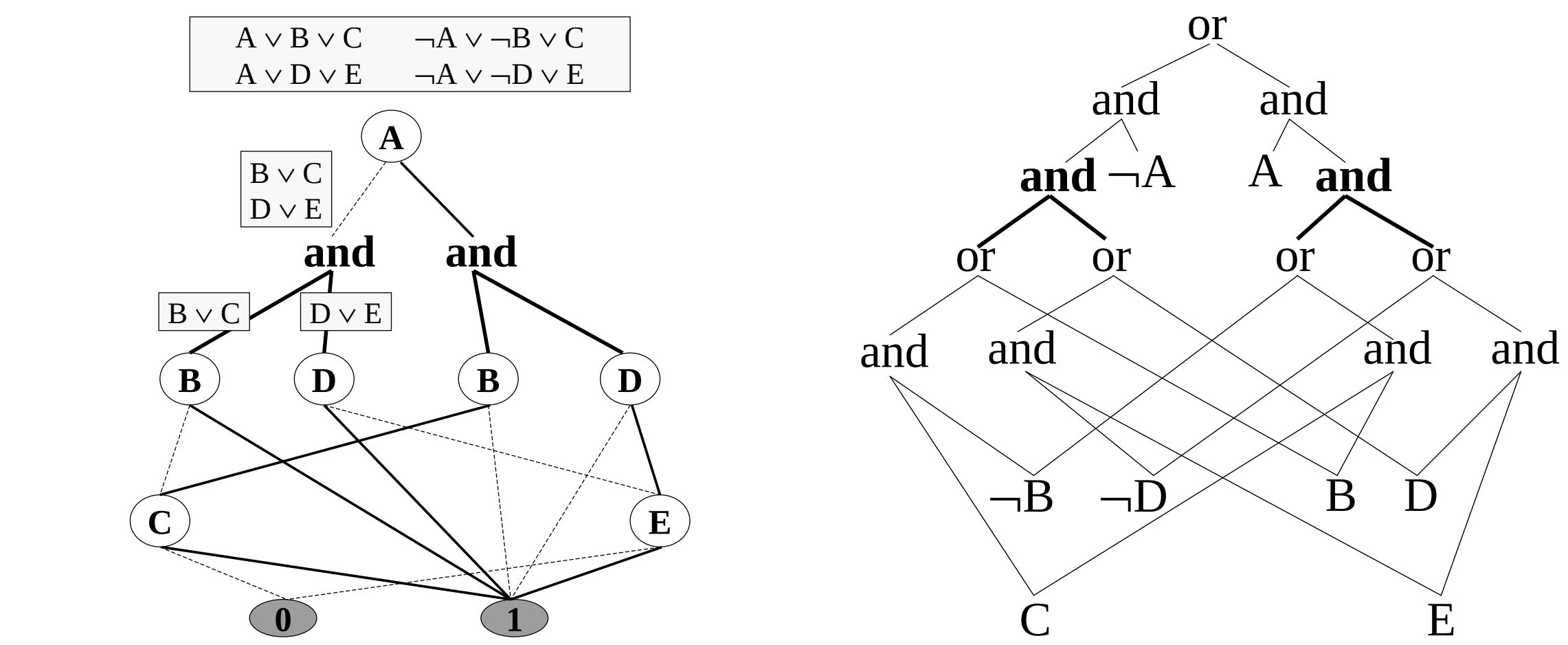

(a) Mixture of decision and conjunction nodes (b) Equivalent NNF circuit

Figure 5: Trace of exhaustive DPLL with decomposition.

comparison of this compilation algorithm and the traditional OBDD construction method can be found in Huang and Darwiche (2005b).

### 4.3 Mapping Exhaustive DPLL with Decomposition to a Subset of d-DNNF

It has been observed, in the particular case of model counting, that the efficiency of exhaustive DPLL can be improved by introducing *decomposition*, also known as *component analysis* (Bayardo & Pehoushek, 2000; Bacchus et al., 2003b; Sang et al., 2004, 2005). The idea is that when a propositional formula breaks down to a conjunction of disjoint subformulas (i.e., those that do not share variables), each subformula can be processed separately and the results combined.

Algorithm 6 implements decomposition in exhaustive DPLL by relaxing a constraint on Algorithm 4: Immediately before Line 8 of Algorithm 4, we need not insist any more that a case analysis be performed on some variable $x$ of the formula; instead, we will examine the current formula, and attempt to decompose it (Line 5) into subsets that do not share a

---

**Algorithm 5** $\text{DPLL}_o$(CNF: $\Delta$, order: $\pi$): exhaustive DPLL with fixed variable ordering

---

**if** there is an empty clause in $\Delta$ **then**
  return 0-sink
**if** there are no variables in $\Delta$ **then**
  return 1-sink
$key$ = compute-key($\Delta$)
**if** there exists a cache entry $(key, result)$ **then**
  return $result$
$x$ **= first variable of order** $\pi$ that appears in $\Delta$
$result$ = get-node($x$, $\text{DPLL}_o(\Delta|_{x=0}, \pi)$, $\text{DPLL}_o(\Delta|_{x=1}, \pi)$)
cache-insert($key$, $result$)
return $result$

---

variable (we assume that this process is nondeterministic; that is, we do not have to detect all decomposition points). The search will then run on each of these subformulas separately and recursively (Lines 7–9), and the separate "subtraces" that result are connected by means of an **and-node** to indicate that the results of the recursive call are being combined (Line 10). In case no decomposition is performed (Line 6 fails), we will branch on a selected variable as in regular DPLL (Lines 14&15).

Figure 5a shows the result of an example execution of this algorithm, where the instantiation of the first variable breaks the CNF formula into two disjoint clauses, which are processed separately and the results combined as an and-node. Figure 5b shows the same trace drawn equivalently as an explicit NNF circuit (for ease of viewing constants have been removed and decision nodes at the bottom compacted into the corresponding literals they represent).

As we have just witnessed, the use of decomposition in exhaustive DPLL has resulted in a new type of node in the trace, returned by **get-and-node** on Line 10 of Algorithm 6. The old get-node function (Line 15) still returns decision nodes (in a relaxed sense, as their children now are not necessarily decision nodes) in the form of Figure 1c. The unique nodes technique can also be extended in a straightforward way so that isomorphic and-nodes, as well as duplicate children of an and-node, will not be created.

We are now ready to discuss the proposition language defined by Algorithm 6, for which purpose we first define the following subset of the d-DNNF language, where determinism is fulfilled by means of decision nodes (again, in the relaxed sense):

**Definition 2.** *The language of decision-DNNF is the set of d-DNNF circuits in which all disjunction nodes have the form of Figure 1b, or* $(x \wedge \alpha) \vee (\neg x \wedge \beta)$*, where* $x$ *is a variable.*[2]

2. Note that, unlike in an FBDD, here $\alpha$ and $\beta$ can be either a conjunction or a disjunction node.

**Algorithm 6** DPLL$_d$(CNF: $\Delta$): exhaustive DPLL with decomposition

1: **if** there is an empty clause in $\Delta$ **then**
2:    return 0-sink
3: **if** there are no variables in $\Delta$ **then**
4:    return 1-sink
5: *components* = exhaustive partitions of $\Delta$ with disjoint variable sets
6: **if** $|components| > 1$ **then**
7:    *conjuncts* = $\{\}$
8:    **for all** $\Delta_c \in components$ **do**
9:       *conjuncts* = *conjuncts* $\cup$ $\{$DPLL$_d(\Delta_c)\}$
10:    return get-and-node(*conjuncts*)
11: *key* = compute-key($\Delta$)
12: **if** there exists a cache entry (*key*, *result*) **then**
13:    return *result*
14: select variable $x$ of $\Delta$
15: *result* = get-node($x$, DPLL$_d(\Delta|_{x=0})$, DPLL$_d(\Delta|_{x=1})$)
16: cache-insert(*key*, *result*)
17: return *result*

We can now formally state our result (again, we assume that circuits are always given in their reduced form by application of appropriate reduction rules as described earlier, although we have allowed redundancy in some of our figures for ease of viewing):

**Theorem 3.** *DAGs returned by Algorithm 6 form the language of (reduced) decision-DNNF.*

We are hence provided with a CNF-to-decision-DNNF compiler in Algorithm 6, which can serve as a d-DNNF compiler in practice since decision-DNNF $\subset$ d-DNNF. This means that once the search finishes, we can answer in polynomial time any query about the input propositional formula, as long as that query is known to be tractable for the language of d-DNNF (see Table 1). On the other hand, Algorithm 6 will not be able to finish in polynomial time on any propositional theory that does not have a polynomial-size representation in d-DNNF (and decision-DNNF), no matter what variable ordering and decomposition method it uses.

Again, one needs to implement some form of formula caching to make Algorithm 6 a practical compiler. Several caching methods have been proposed for d-DNNF compilation, the latest and most effective of which appeared in Darwiche (2004). However, we refer the reader to Darwiche (2001) for a caching scheme that is specific to a decomposition method based on what is known as dtrees (which we discuss next). This scheme is not as effective as the one in Darwiche (2004) in that the former may miss some equivalences that would be caught by the latter, yet it allows one to show that the space and time complexity of Algorithm 6, with this caching scheme in force, will be exponential only in the *treewidth* of the CNF formula (as compared to pathwidth and cutwidth in OBDD compilation as discussed in Section 4.2). Considering that model counting is a linear-time query supported by the d-DNNF language, the results of Darwiche (2001) also imply that DPLL with decomposition (such as Algorithm 6) can be used to count models with a time and space complexity that is exponential only in the treewidth of the CNF formula; see Bacchus et al. (2003a) for an alternative derivation of this complexity result. Interestingly, no similar structure-based measure of complexity appears to be known for FBDD compilation.

Finally, we would like to briefly discuss a distinction between two possible methods of decomposition. Algorithm 6 suggests a dynamic notion of decomposition, where disjoint components will be recognized after each variable split. This dynamic decomposition was initially proposed and utilized by Bayardo and Pehoushek (2000) for model counting and adopted by more recent model counters (Sang et al., 2004, 2005). Darwiche (2002, 2004) proposed another method for performing the decomposition less dynamically by preprocessing the CNF formula to generate a *dtree (decomposition tree)*, which is a binary tree whose leaves correspond to clauses of the CNF formula. Each node in the dtree defines a set of variables, called the *cutset,* whose instantiation is guaranteed to decompose the CNF formula below that node into disjoint components. The rationale is that the cost of dynamically computing a partition (Line 5 of Algorithm 6) many times during the search is now replaced with the lesser cost of computing a static and recursive partition once and for all. This method of decomposition allows one to provide structure-based computational guarantees as discussed above. Moreover, the instantiation of variables in each cutset can be performed dynamically, by utilizing dynamic variable ordering heuristics as typically done in SAT solvers. The use of dtrees, combined with dynamic variable ordering, leads to an almost static behavior on highly structured problems, for which cutsets are small. Yet, one sees a more dynamic behavior on less structured problems, such as random 3-SAT, where the cutsets are relatively large and dynamic variable ordering tends to dominate. Interestingly, the static behavior of dtrees (low overhead) can be orders of magnitude more efficient than purely dynamic behavior on structured benchmarks, including the ISCAS85 circuits. On the other hand, the dynamic behavior of dtrees can lead to very competitive results on unstructured benchmarks, including random 3-SAT. One may obtain results to this effect by running the model counter of Sang et al. (2004), Cachet Version 1.1, against the d-DNNF compiler of Darwiche (2004), c2d Version 2.2, on relevant benchmarks. It should be noted that the two programs differ in other aspects, but the decomposition method appears to be one of the major differences.

### 4.4 Harnessing Search Techniques for Knowledge Compilation

Research in recent years has greatly improved the efficiency and scalability of systematic search methods, particularly those for the problem of propositional satisfiability. Techniques contributing to this improvement include sophisticated conflict analysis, dependency-directed backtracking, clause learning, new variable ordering heuristics, and data structures for faster constraint propagation, among other things (Marques-Silva & Sakallah, 1996; Marques-Silva, 1999; Aloul, Markov, & Sakallah, 2001; Moskewicz, Madigan, Zhao, Zhang, & Malik, 2001; Zhang, Madigan, Moskewicz, & Malik, 2001; Goldberg & Novikov, 2002; Zhang & Malik, 2002; Heule & van Maaren, 2004). On the other hand, we have described in this section a uniform framework where systematic search algorithms can be converted into knowledge compilers by exhausting the search space and recording the trace of the search. Such a framework affords an opportunity for many of the successful search techniques to carry over to knowledge compilation. We refer the reader to Bayardo and Pehoushek (2000), Darwiche (2002, 2004), Huang and Darwiche (2005b), and Sang et al. (2004, 2005) for detailed discussions of issues that arise from the implementation of these techniques when search is extended to exhaustion, when the trace of the search needs to be stored, and when decomposition is introduced.

Finally, we note that the efficiency of the search as addressed by these and other techniques (including caching in particular) is an important practical issue, but is orthogonal to the language generated by the search, the main focus of the present paper. As a simple example, one may have two versions of Algorithm 5 that have drastically different running times on the same input because one of them has a much better caching method, but at the end of the day, they are bound to produce exactly the same OBDD due to the canonicity of OBDDs. As another example, a learned clause provided by conflict analysis can reduce the number of search nodes, but of itself will not affect the final DAG (circuit) generated, because the nodes avoided all correspond to contradictions (the constant 0) and would not appear in the DAG anyway due to the reduction rules.

## 5. Power and Limitations of Exhaustive Search Algorithms

In Sections 3 and 4 we have established the notion of interpreting the trace of an exhaustive search as a circuit, and then mapping the search algorithm itself to a propositional language consisting of all its possible traces. This notion provides a new perspective on the intrinsic power and limitations of these search algorithms, which we have illustrated in Section 4 by discussing the usefulness of the search algorithms as knowledge compilers and the inherent inefficiency of the same on certain classes of inputs. In this section we formalize these concepts and further illustrate them using examples from real implementations of exhaustive search algorithms.

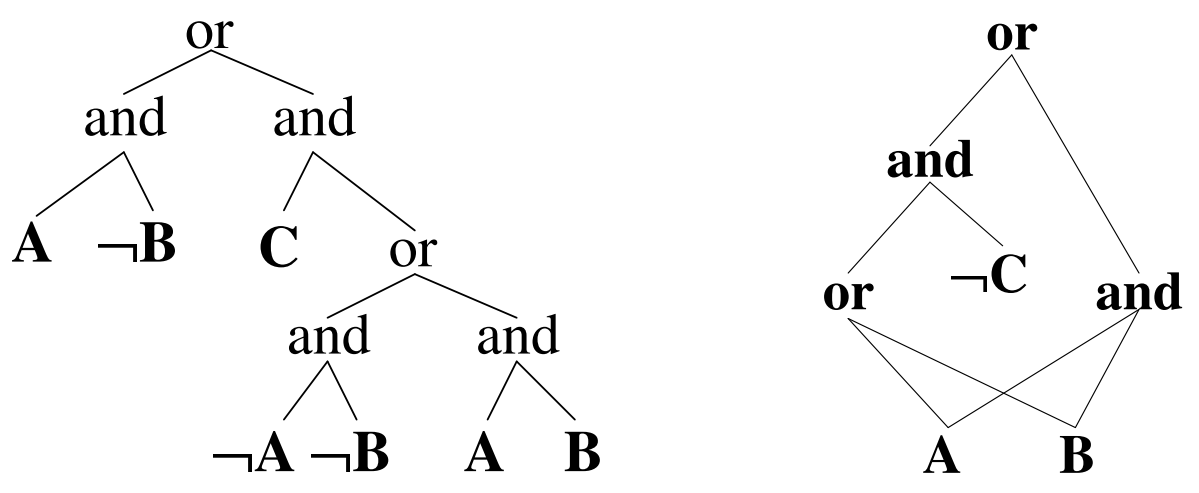

(a) General determinism (b) Nondeterminism

Figure 6: DPLL is unable to produce general determinism or nondeterminism.

Consider an arbitrary exhaustive search algorithm based on variable splitting, call it $\text{DPLL}_x$, and suppose that its traces form a propositional language $L_x$. The intrinsic power and limitations of $\text{DPLL}_x$ can then be identified by the following two principles, respectively:

1. If $\text{DPLL}_x$ runs in polynomial time on a class of formulas, then $\text{DPLL}_x$ (with its trace recorded) can answer in polynomial time any query on these formulas that is known to be tractable for language $L_x$.

2. $\text{DPLL}_x$ will not run in polynomial time on formulas for which no polynomial-size representations exist in $L_x$.

Take for example the model counters recently proposed by Bayardo and Pehoushek (2000) and Sang et al. (2004, 2005), which employ the techniques of decomposition and (the latter also) formula caching. A simple analysis of these model counters shows that their traces are in the language of decision-DNNF.[3] Now consider the query of testing whether the minimization of a theory $\Delta$ implies a particular clause $\alpha$, $\min(\Delta) \models \alpha$, where $\min(\Delta)$ is defined as a theory whose models are exactly the minimum-cardinality models of $\Delta$. This query is at the heart of diagnostic and nonmonotonic reasoning and is known to be tractable if $\Delta$ is in d-DNNF. Applying the first principle above, and noting that decision-DNNF $\subset$ d-DNNF, we conclude that this query can be answered in polynomial time for any class of formulas on which the model counters of Bayardo and Pehoushek (2000) and Sang et al. (2004, 2005) have a polynomial time complexity. Similarly, a probabilistic equivalence test can be performed in polynomial time for formulas on which these models counters have a polynomial time complexity.

As an example of the second principle above, first note that neither of these same model counters will finish in polynomial time on formulas for which no polynomial-size representations exist in decision-DNNF. Furthermore, recall that decision-DNNF as defined in Definition 2 is a strict subset of d-DNNF: Every disjunction in a decision-DNNF circuit

3. See the DDP algorithm of Bayardo and Pehoushek (2000) and Table 1 of Sang et al. (2004). For example, the variable splitting (Lines 5–9) in Table 1 of Sang et al. (2004) corresponds to the generation of a decision node, and the ToComponents function (Line 6 and Line 8) corresponds to the generation of an and-node that satisfies decomposability.

has the form $(x \wedge \alpha) \vee (\neg x \wedge \beta)$, while d-DNNF allows disjunctions of the form $\eta \vee \psi$ where $\eta \wedge \psi$ is logically inconsistent, yet $\eta$ and $\psi$ do not contradict each other on any particular variable (Figure 6a gives one example). Recall also that the model counting query remains tractable when one generalizes from decision-DNNF to d-DNNF. If decision-DNNF turns out to be not as succinct as d-DNNF, therefore, one may find another generation of model counters, as well as d-DNNF compilers that can be exponentially more efficient than the current ones.

Finally, we note that DPLL traces are inherently bound to be NNF circuits that are both deterministic and decomposable. Decomposability alone, however, is sufficient for the tractability of such important tasks as clausal entailment testing, existential quantification of variables, and cardinality-based minimization (Darwiche & Marquis, 2002). DPLL cannot generate traces in DNNF that are not in d-DNNF (Figure 6b for example), since variable splitting (the heart of DPLL) amounts to enforcing determinism. It is the property of determinism that provides the power needed to do model counting (#SAT), which is essential for applications such as probabilistic reasoning. But if one does not need this power, then one should go beyond DPLL-based procedures; otherwise one would be solving a harder computational problem than is necessary.

## 6. Relation to Previous Work

We have employed the notion of the *trace* of a search in this paper to provide a theoretical and practical channel between advances in systematic search and those in knowledge compilation. This channel has indeed been active for some time, but implicitly and mostly in one direction: systematic search algorithms being employed to compile knowledge bases (see, for example, Darwiche, 2002, 2004; Huang & Darwiche, 2005b; Darwiche, 2005). In fact, the techniques of variable ordering, decomposition, and caching have all been extensively used in these bodies of work, just as they are being used (some more recently) in pure search algorithms.

A key contribution of this paper is then in formally explicating this notion of search trace, and proposing it as the basis for a systematic framework for *compiling* knowledge bases into subsets of NNF. This is to be contrasted with earlier systematic studies on NNF (Darwiche & Marquis, 2002), as those were concerned with *describing* the various properties of compiled NNF representations without delving into the algorithmic nature of generating them. Another key contribution of this paper is in activating the second direction of the search/compilation channel: looking at the language membership of search traces to formally characterize the power and limitations of various search algorithms.

There has been a recent line of work, centering on the notion of *AND/OR search,* which is also concerned with understanding the power and limitations of various search techniques, such as decomposition and caching, by measuring the size of explored search spaces (Dechter & Mateescu, 2004b, 2004a; Marinescu & Dechter, 2005). The premise of this work is that while traditional search algorithms (based on branching) can be thought of as exploring an OR–search space, more recent search algorithms (employing decomposition) can be thought of as exploring an AND/OR–search space. Here, the space of an AND/OR search is characterized by a graph (or tree if no caching is used) with alternating layers of AND-nodes and OR-nodes, the former representing decomposition and the latter branching. The algorithms for exploring an AND/OR–search space therefore exhibit a behavior similar to that of Algorithm 6, used in compiling CNF to d-DNNF (Darwiche, 2002, 2004, 2005)), except that Algorithm 6 records the AND/OR–search space explicitly as a circuit. Hence, AND/OR search and Algorithm 6 share the same limitations we discussed of DPLL: They cannot take advantage of the general notion of determinism, or rid themselves of determinism altogether.

We note here that the proposed algorithms for AND/OR search proceed by instantiating variables and performing decomposition according to a *pseudo tree* (Freuder & Quinn, 1985), while earlier work on d-DNNF compilation uses a *decomposition tree* for those two tasks (corresponding to the choices on Lines 5 and 14 of Algorithm 6). Pseudo trees and decomposition trees are similar in that they both provide a scheme in which the instantiation of a certain set of variables will lead to decomposition of the problem. In the framework proposed in this paper, however, we make no commitment to either the decomposition or the variable ordering scheme, as they are not relevant to our discussion. One should note though that a commitment to any particular decomposition or variable ordering scheme can have significant practical implications. Specifically, a search algorithm whose trace is in d-DNNF may be performing variable ordering and decomposition in such a way as to prohibit the possibility of generating certain (space efficient) d-DNNF circuits.

Finally, one can identify a major difference between AND/OR search and d-DNNF compilation, in terms of their handling of queries: Each execution of an AND/OR search algorithm answers only a single query, while executing a d-DNNF compilation algorithm results in a compact structure that can be used repeatedly to answer all queries (for the same knowledge base) that are known to be tractable. As discussed earlier, traversing a compiled structure can be potentially much more efficient than repeating the search that produced it. From this point of view, the separation of search from the actual reasoning task provides the benefit of amortizing the search effort over a potentially large number of queries. It also provides, as we have discussed, a systematic methodology by which *independent* advances in search can be harnessed to improve the performance of automated reasoning systems.

## 7. Experimental Results

By way of experimentation, we ran implementations of Algorithm 5 using the MINCE variable ordering heuristic (Aloul et al., 2001), Algorithm 4 using the VSIDS variable ordering heuristic (Moskewicz et al., 2001), and Algorithm 6 using static decomposition by hypergraph partitioning (Darwiche & Hopkins, 2001), to compile a set of CNF formulas into OBDD, FBDD, and d-DNNF, respectively (implementation details of the first and third programs can be found in Huang & Darwiche, 2005b and Darwiche, 2004). The goal of these experiments is to show the practicality of the search-based compilation framework and to illustrate the improvement of language succinctness in response to the relaxation of constraints on the search process. The benchmarks we used include random 3-CNF and graph coloring problems from Hoos and Stützle (2000), and a set of ISCAS89 circuits.

The results of these experiments are shown in Table 3, where the running times are given in seconds based on a 2.4GHz CPU. The size of the compilation reflects the number of edges in the NNF DAG. A dash indicates that the compilation did not succeed given the available memory (4GB) and a 900-second time limit. It can be seen that for most of these instances, the compilation was the smallest in d-DNNF, then FBDD, then OBDD; a similar relation can be observed among the running times. Also, the number of instances successfully compiled was the largest for d-DNNF, then FBDD, then OBDD. This tracks well with the theoretical succinctness relations of the three languages. (However, note that FBDD and d-DNNF are not canonical representations and therefore compilations smaller than reported here are perfectly possible; smaller OBDD compilations are, of course, also possible under different variable orderings.)

Table 3: Compiling CNF into OBDD, FBDD, and d-DNNF.

| CNF Name | Number of Models | OBDD Size | OBDD Time | FBDD Size | FBDD Time | d-DNNF Size | d-DNNF Time |
|---|---|---|---|---|---|---|---|
| uf75-01 | 2258 | 10320 | 0.14 | 3684 | 0.02 | 822 | 0.02 |
| uf75-02 | 4622 | 22476 | 0.15 | 14778 | 0.04 | 1523 | 0.03 |
| uf75-03 | 3 | 450 | 0.02 | 450 | 0.02 | 79 | 0.01 |
| uf100-01 | 314 | 2886 | 2.22 | 2268 | 0.01 | 413 | 0.02 |
| uf100-02 | 196 | 1554 | 0.91 | 1164 | 0.07 | 210 | 0.04 |
| uf100-03 | 7064 | 12462 | 0.78 | 9924 | 0.12 | 1363 | 0.02 |
| uf200-01 | 112896 | 8364 | 651.04 | 7182 | 35.93 | 262 | 3.66 |
| uf200-02 | 1555776 | – | – | 12900 | 33.72 | 744 | 2.64 |
| uf200-03 | 804085558 | – | – | 662238 | 56.61 | 86696 | 10.64 |
| flat75-1 | 24960 | 23784 | 0.16 | 10758 | 0.04 | 2273 | 0.01 |
| flat75-2 | 774144 | 13374 | 0.28 | 8844 | 0.04 | 1838 | 0.01 |
| flat75-3 | 25920 | 84330 | 0.29 | 26472 | 0.07 | 4184 | 0.04 |
| flat100-1 | 684288 | 62298 | 0.78 | 37704 | 0.10 | 3475 | 0.03 |
| flat100-2 | 245376 | 88824 | 1.57 | 39882 | 0.30 | 6554 | 0.09 |
| flat100-3 | 11197440 | 15486 | 0.15 | 21072 | 0.09 | 2385 | 0.02 |
| flat200-1 | 5379314835456 | – | – | – | – | 184781 | 56.86 |
| flat200-2 | 13670940672 | – | – | 134184 | 7.07 | 9859 | 23.81 |
| flat200-3 | 15219560448 | – | – | 358092 | 4.13 | 9269 | 3.28 |
| s820 | 8388608 | 1372536 | 72.99 | 364698 | 0.69 | 23347 | 0.07 |
| s832 | 8388608 | 1147272 | 76.55 | 362520 | 0.70 | 21395 | 0.05 |
| s838.1 | 73786976294838206464 | 87552 | 0.24 | – | – | 12148 | 0.02 |
| s953 | 35184372088832 | 2629464 | 38.81 | 1954752 | 4.01 | 85218 | 0.26 |
| s1196 | 4294967296 | 4330656 | 78.26 | 4407768 | 12.49 | 206830 | 0.44 |
| s1238 | 4294967296 | 3181302 | 158.84 | 4375122 | 12.14 | 293457 | 0.94 |
| s1423 | 2475880078570760549798248448 | – | – | – | – | 738691 | 4.75 |
| s1488 | 16384 | 6188427 | 50.35 | 388026 | 1.14 | 51883 | 0.19 |
| s1494 | 16384 | 3974256 | 31.67 | 374760 | 1.07 | 55655 | 0.18 |

We close this section by noting that the implementations of these knowledge compilers bear witness to the advantage of the search-based compilation framework we have described in Section 4. The first compiler is based on an existing SAT solver (Moskewicz et al., 2001), and the other two on our own implementation of DPLL, all three benefiting from techniques that have found success in SAT, including conflict analysis, clause learning, and data structures for efficient detection of unit clauses.

## 8. Conclusion

This work is concerned with a class of exhaustive search algorithms that are run on propositional knowledge bases. We proposed a novel methodology whereby the trace of a search is identified with a combinational circuit and the search algorithm itself is mapped to a propositional language consisting of all its possible traces. This mapping leads a uniform and practical framework for compilation of propositional knowledge bases into various languages of interest, and at the same time provides a new perspective on the intrinsic power and limitations of exhaustive search algorithms. As interesting examples, we unveiled the "hidden power" of several recent model counters, discussed one of their potential limitations, and pointed out the inability of this class of algorithms to produce traces without the property of determinism, which limits their power from a knowledge compilation point of view. We discussed the generality of some of of our results in relation to recent work on AND/OR search. Finally, we presented experimental results to demonstrate the practicality of the search-based knowledge compilation framework and to illustrate the variation of language succinctness in response to the variation of the search strategy.

## Acknowledgments

Parts of this work have been presented in "DPLL with a Trace: From SAT to Knowledge Compilation," in *Proceedings of the* $19^{th}$ *International Joint Conference on Artificial Intelligence (IJCAI)*, 2005, pages 156–162. We thank Rina Dechter and the anonymous reviewers for their feedback on earlier drafts of this paper. This work was partially supported by NSF grant IIS-9988543, MURI grant N00014-00-1-0617, and JPL/NASA contract 442511-DA-57765. National ICT Australia is funded by the Australian Government's *Backing Australia's Ability* initiative, in part through the Australian Research Council.

## Appendix A

### Proof of Theorem 1

We first point out that the recursion is guaranteed to terminate as each recursive call (Line 6) is accompanied by the disappearance of one variable and eventually either Line 2 or Line 4 will execute. Assuming the compact drawing of decision nodes as in Figure 1c, we now show that every DAG returned by Algorithm 2 is an FBDD, and that every FBDD can be generated by some execution of Algorithm 2.

Part I: There are three return statements in Algorithm 2: Lines 2, 4, and 6. The single nodes returned on Lines 2 and 4 are trivial FBDDs. The DAG returned by get-node on Line 6, call it $G$, is a decision node by induction on the level of recursion. Hence it remains to show that $G$ satisfies the test-once property (equivalent to decomposability in this case as we discussed in Section 2.1), which is true because on Line 6, variable $x$ has been replaced with constants before the two recursive calls, and hence cannot appear in the two graphs that are supplied as the second and third arguments to get-node. Finally, FBDDs returned by the algorithm are guaranteed to be reduced by the use of the unique nodes technique in the get-node function (Brace et al., 1991).

Part II: Let $G$ denote an arbitrary (reduced) FBDD, and also its root node. Define the following function $\Delta(G)$ that returns a CNF formula (as a set of clauses) for every $G$:

$$\Delta(G) \equiv \begin{cases} \{\text{the empty clause}\}, & \text{if } G \text{ is the 0-sink;} \\ \{\}, & \text{if } G \text{ is the 1-sink;} \\ \{x \vee c \mid c \in \Delta(G.low)\} \cup \{\neg x \vee c \mid c \in \Delta(G.high)\}, & \text{otherwise, where node } G \\ & \text{is labeled with variable } x. \end{cases}$$

On the last line of the definition above, assume that the literal $x$ (and $\neg x$) is always appended to the front of the clause $c$. Now execute Algorithm 2 on $\Delta(G)$, and on Line 5 always choose the first variable of any clause (the first variable of every clause must be the same). The DAG returned will be isomorphic to $G$. □

**Proof of Theorem 2**

We show that every DAG returned by Algorithm 5 is an OBDD, and that every OBDD can be generated by some execution of Algorithm 5.

Part I: First, any DAG returned by Algorithm 5 is an FBDD as Algorithm 5 is a restricted version of Algorithm 2. Second, the DAG is an OBDD because any nonterminal node $N$ in the DAG must be constructed on Line 9, and therefore by virtue of Line 8 satisfies the following property: The variable $x$ that labels $N$ appears before all other variables in the two subgraphs below $N$ in the variable order $\pi$. Finally, OBDDs returned by the algorithm are guaranteed to be reduced by the use of the unique nodes technique in the get-node function (Brace et al., 1991).

Part II: Given an arbitrary (reduced) OBDD $G$, let $\pi$ be the variable order of $G$. Now using $\Delta(G)$ as defined in the previous proof, execute Algorithm 5 on $(\Delta(G), \pi)$. The DAG returned will be isomorphic to $G$. □

**Proof of Theorem 3**

We show that every DAG returned by Algorithm 6 is a decision-DNNF circuit, and that every decision-DNNF circuit can be generated by some execution of Algorithm 6.

Part I: Nodes returned by get-and-node on Line 10 are conjunction nodes that satisfy decomposability because the components identified on Line 5 do not share variables. Nodes returned by get-node on Line 15 are disjunction nodes that have the form of Figure 1c by the definition of get-node. Therefore the whole DAG is in decision-DNNF. Finally, decision-DNNF circuits returned by the algorithm are guaranteed to be reduced by the use of the unique nodes technique in the get-node and get-and-node functions.

Part II: Let $G$ denote an arbitrary (reduced) decision-DNNF circuit. Again assuming the compact drawing of decision nodes as in Figure 1c, we now expand the previous definition of $\Delta(G)$ as follows:

$$\Delta(G) \equiv \begin{cases} \{\text{the empty clause}\}, & \text{if } G \text{ is the 0-sink;} \\ \{\}, & \text{if } G \text{ is the 1-sink;} \\ \bigcup_i \Delta(G_i), & \text{if } G \text{ is } \bigwedge_i G_i; \\ \{x \vee c \mid c \in \Delta(G.low)\} \cup \{\neg x \vee c \mid c \in \Delta(G.high)\}, & \text{otherwise, where node } G \\ & \text{is labeled with variable } x. \end{cases}$$

As in the previous definition of $\Delta(G)$, assume that on the last line of the definition above, the literal $x$ (and $\neg x$) is always appended to the front of the clause $c$. Now, let each literal of each clause be associated with a (possibly empty) list of "colors" as follows: The lists are initially empty for the literals $x$ and $\neg x$ introduced on the last line of the definition; on the third line of the definition, assign a distinct color to each of the sets being unioned, and in each set, append the assigned color to the head of the list of colors for the first literal of every clause. Now execute Algorithm 6 on $\Delta(G)$ and resolve the nondeterministic choices on Line 5 (decomposition) and Line 14 (variable selection) as follows: If the first literal of any clause has a nonempty list of colors, then (the first literal of every clause must have a nonempty list of colors) let Line 5 partition the set of clauses according to the first color of their first literal and remove that color from the first literal of every clause; otherwise, (the first literal of every clause must mention the same variable) let Line 5 return a single partition and on Line 14 choose the first variable of any clause. The DAG returned will be isomorphic to $G$. $\square$